\pdfoutput=1

\documentclass[runningheads]{llncs}

\usepackage{times}
\usepackage{latexsym}
\usepackage[T1]{fontenc}
\usepackage[utf8]{inputenc}
\usepackage{microtype}
\usepackage{inconsolata}

\usepackage{multirow}
\usepackage{tabularray}
\usepackage{graphicx}
\usepackage{todonotes}
\usepackage{amsmath}
\usepackage{array}
\usepackage{enumitem}
\usepackage{hyperref}
\usepackage{algorithm}
\usepackage{algpseudocode}
\usepackage[most]{tcolorbox}
\usepackage{color, colortbl}
\definecolor{Gray}{gray}{0.9}
\usepackage{arydshln}
\usepackage{subcaption}
\usepackage{subcaption}
\usepackage{cleveref}

\begin{document}

\title{Sparse Graph Representations for Procedural Instructional Documents}

\author{Shruti Singh$^*$\inst{1} \and Rishabh Gupta\inst{2}}
\authorrunning{Singh and Gupta}
\institute{Indian Institute of Technology Gandhinagar, India \and
Bosch Research and Technology Centre, Bangalore, India \\
\email{singh\_shruti@iitgn.ac.in, gupta.rishabh@in.bosch.com}\\
}

\maketitle  

\def\thefootnote{*}\footnotetext{work done during internship at Bosch}\def\thefootnote{\arabic{footnote}}

\begin{abstract}
Computation of document similarity is a critical task in various NLP domains that has applications in deduplication, matching, and recommendation. 
Graph representations such as Joint Concept Interaction Graph (JCIG) represent a pair of documents as a joint undirected weighted graph. 
JCIGs facilitate an interpretable representation of document pairs as a graph. However, JCIGs are undirected, and don’t consider the sequential flow of sentences in documents. We propose two approaches to model document similarity by representing document pairs as a directed and sparse JCIG that incorporates sequential information. We propose two algorithms inspired by Supergenome Sorting and Hamiltonian Path that replace the undirected edges with directed edges. Our approach also sparsifies the graph to $O(n)$ edges from JCIG's worst case of $O(n^2)$. We show that our sparse directed graph model architecture consisting of a Siamese encoder and GCN achieves comparable results to the baseline on datasets not containing sequential information and beats the baseline by ten points on an instructional documents dataset containing sequential information.

\keywords{Document Similarity \and Concept Interaction Graphs \and GCN}
\end{abstract}
\section{Introduction and Related Works}
Document similarity has applications in detecting duplicate documents collected from diverse sources, paraphrase detection, and recommendation of similar articles. Several datasets such as quora question pairs~\cite{qqpdata}, microsoft research paraphrase corpus~\cite{dolan-brockett-2005-automatically}, STS benchmark~\cite{cer-etal-2017-semeval} for sentence similarity exist that address the task of semantic similarity computation. Recently, all these dataset leaderboards are dominated by transformer-based models~\cite{vaswani2017attention}. However, all these datasets contain short documents (few sentences only) and transformer-based models don't scale quadratically to long documents. 

Joint concept interaction graphs (JCIG) proposed by~\cite{liu-etal-2019-matching} represents a pair of documents as a joint undirected weighted graph, where nodes of the graph represent the concepts in the documents, and the edges represent the interaction between the concepts. JCIGs show state-of-the-art performance in detecting similar chinese news articles, also beating a fine-tuned BERT baseline. However, JCIGs don't encode sequential information in documents which is crucial for certain domains such as recipes, instruction manuals, and compliance documents. In such domains, sequence of steps is crucial, and ignoring the order and following steps randomly can lead to damage of devices or undesirable end products. Hence the need to encode the sequential information while matching documents for such domains is important. Several procedural text datasets have been curated such as RecipeQA~\cite{yagcioglu-etal-2018-recipeqa} consisting of recipe documents, EMQAP~\cite{nandy-etal-2021-question-answering} of emanual documents, and ProPara~\cite{dalvi-etal-2018-tracking} designed to track changing entity states in procedural text. However, all these works model the task in question-answering format, while our work focuses on the document similarity task. In this work, we focus on JCIG~\cite{liu-etal-2019-matching} and propose improvements over it to encode sequential information.

\begin{figure*}[htbp]
    \centering
    \small{
        \begin{subfigure}[b]{0.95\textwidth}
             \centering
             \includegraphics[width=0.95\textwidth]{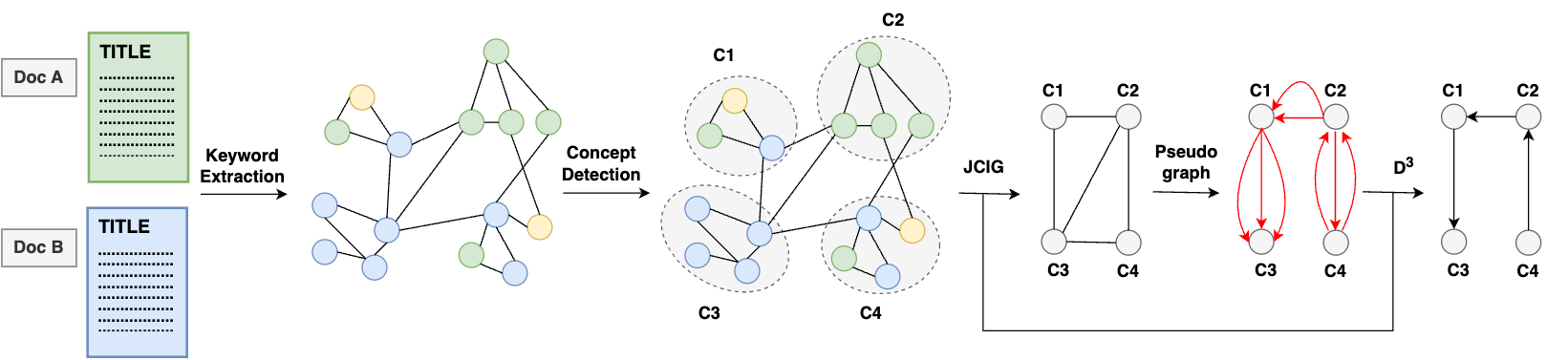}
             \caption{Directed JCIG construction from document pair.}
             \label{fig:jcigconstruction}
         \end{subfigure}
         \begin{subfigure}[b]{0.9\textwidth}
             \centering
             \includegraphics[width=0.9\textwidth]{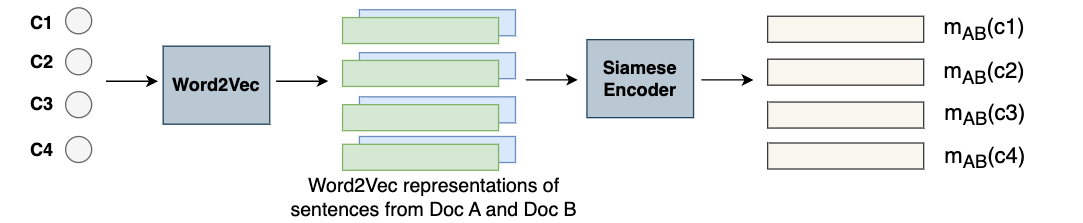}
             \caption{Siamese encoder for node representations.}
             \label{fig:siamenc}
         \end{subfigure}
         \begin{subfigure}[b]{0.7\textwidth}
             \centering
             \includegraphics[width=0.7\textwidth]{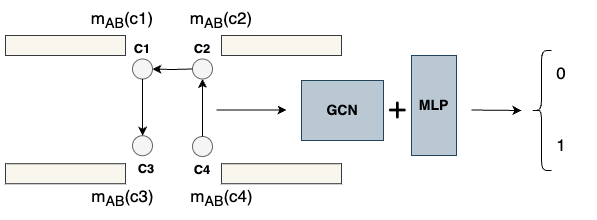}
             \caption{GCN encoder and MLP classifier.}
             \label{fig:gcnmlp}
         \end{subfigure}
    }
    \caption{Overall pipeline of the document similarity architecture that has three major steps: (a) Directed JCIG construction, (b) Siamese encoder, and (c) GCN encoder and MLP classifier.}
    \label{fig:neural_architecture}
\end{figure*}

We propose to add directions to JCIGs that encode the sequential information and then sparsify the graph by deducing the dominant direction ($D^3$) of information flow between a pair of nodes. We propose two algorithms for $D^3$: Supergenome Sorting (SGS) and Hamiltonian Path (HP) and showcase that our model beats the JCIG baseline by 10 points on the \textsc{iFixit} dataset which consists of instruction manuals for appliances. For the Chinese news articles datasets (CNSE and CNSS), the sequential information is not crucial, and hence we don't observe significant improvements but also don't observe any drop in performance. The paper is organized as follows. We discuss JCIG construction in~\Cref{sec:method}. We describe the importance of encoding sequential information in~\Cref{sec:encseqinfor}, direction pseudograph in~\Cref{sec:dirpseudograph}, and SGS and HP algorithms in~\Cref{sec:ddd}. This is followed by a description of datasets and experiments in~\Cref{sec:data_exp} and results in~\Cref{sec:results}. A schematic representation of our architecture is presented in~\Cref{fig:neural_architecture}.

\section{Joint Concept Interaction Graph Representation}
\label{sec:method}
We briefly discuss the Joint Concept Interaction Graph (JCIG)~\cite{liu-etal-2019-matching} for representing a pair of documents along with our ablation studies, as it forms the fundamental basis on which we extend in our work. We begin by discussing the construction of a concept interaction graph for a single document in~\Cref{ssec:cig_met}, followed by the construction of a joint concept interaction graph (JCIG) for a pair of documents in~\Cref{ssec:jcig_met}. 
\subsection{Concept Interaction Graph}
\label{ssec:cig_met}
A Concept Interaction Graph (CIG) represents a document $\mathcal{D}_A$ as an undirected weighted graph $\mathcal{G}_{A}(V, E)$, where $V$ is the vertex set and $E$ is the edge set. The vertex set consists of keywords $\mathcal{K}_A$ extracted from document $\mathcal{D}_A$. Next, we describe the primary steps in the construction of a CIG.
\begin{table*}[!bhtp]
    \begin{tabular}{c}
    \includegraphics[width=0.4\linewidth]{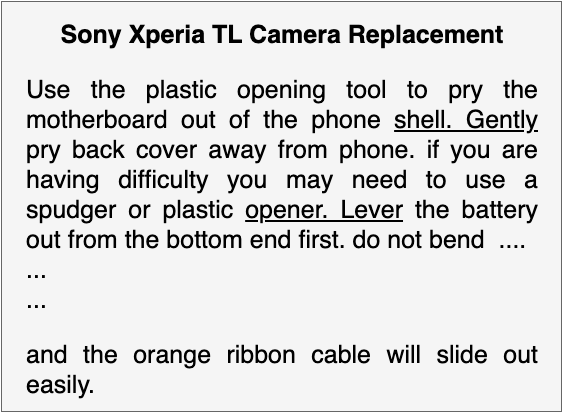}\\
    \end{tabular}
    \quad
    \small{
        \begin{tabular}{p{0.12\linewidth}|p{0.14\linewidth}|p{0.12\linewidth}|p{0.12\linewidth}}
        \rowcolor{Gray}
            \hline \textbf{Summa} & \textbf{RAKE} & \textbf{TextRank} & \textbf{KeyBERT} \\ \hline
             shell gently & camera using tweezers & difficulty & cable gently \\ \hline
             opener lever & gently pry back cover away & spudger or plastic opener & motherboard away \\  \hline
             orange & probably need tweezers & gold & screwdriver gently\\ \hline
        \end{tabular}
    }
    \caption{For a sample document on `camera replacement' from the \textsc{iFixit} dataset, we present flawed keywords extracted by Summa, RAKE, TextRank, and KeyBERT. Based on manual inspection of keywords extracted by each algorithm, we went ahead with the TextRank implementation by~\cite{PyTextRank}. The document on the left highlights the keywords extracted by Summa that are spread across consecutive sentences (marked with underline), which are not optimal keywords.}
    \label{tab:keyword_ext_algos}
\end{table*}
\subsubsection*{\textbf{Keyword Extraction}}
Keywords are extracted from document $\mathcal{D}_A$, using popular keyword extraction algorithms: RAKE~\cite{doi:https://doi.org/10.1002/9780470689646.ch1}, Summa~\cite{barrios2016variations}, TextRank~\cite{mihalcea-tarau-2004-textrank,PyTextRank}, KeyBERT~\cite{grootendorst2020keybert}. Let the keywords extracted in this stage for a document $\mathcal{D}_A$ be denoted by $\mathcal{K}_A = \{kw_1, kw_2, ..., kw_n\}$. \\

TextRank~\cite{mihalcea-tarau-2004-textrank,PyTextRank} constructs an undirected unweighted graph whose nodes are tokens in the text (tokens filtered via syntactic rules to remove stopwords), and edges are added between two nodes if they co-occur within a window of N tokens. PageRank~\cite{BRIN1998107} is used to compute the score for each vertex, and top-k vertices are selected as keywords. KeyBERT~\cite{grootendorst2020keybert} computes the document representation and n-gram phrase representations of n-grams in the document via BERT~\cite{devlin-etal-2019-bert}. It computes the cosine similarity between the document and n-gram phrase representations and selects the most similar n-grams as keywords. KeyBERT requires the range of n-grams to be considered and stopwords to be ignored.
We manually inspected the keywords extracted with RAKE, Summa, TextRank, and KeyBERT for different configurations and observed that the most optimal keywords are extracted via Summa. KeyBERT selects verb phrases, RAKE has a preference for long phrases (usually 3 or 4-grams). Summa performs similarly to TextRank with SpaCy backend, but Summa has some unsuitable phrases which are spread across consecutive sentences. We present a sample of flawed keywords extracted via each algorithm in Table~\ref{tab:keyword_ext_algos}. Based on manual analysis of keywords from multiple documents, we employed TextRank implementation by~\cite{PyTextRank}.

\subsubsection*{\textbf{Keyword Graph ($\mathcal{KG}$)}} After extraction of keywords in a document is Keyword Graph ($\mathcal{KG}$) construction. The vertices of the Keyword Graph for the document $\mathcal{D}_A$ are initialized from the keyword set $\mathcal{K}_A$. An edge is added between keywords $kw_i$ and $kw_j$ if they co-occur in the same sentence. Let $S_A$ be the set of sentences in the document $\mathcal{D_A}$, and each sentence $s_i \in S_A$ is a set of words and keywords. $\mathcal{KG}_{A}(V)$ and $\mathcal{KG}_{A}(E)$ denote the vertex set and edge set of $\mathcal{KG}_{A}$.
\begin{eqnarray*}
S_A &=& [s_1, s_2, ..., s_m] \\
s_i &=& \{w_1, w_2, ..., kw_j, kw_k, ..\}, \\
& & s_i \in S_A \\
\mathcal{KG}_{A}(V) &=& \mathcal{K}_A = \{kw_1, kw_2, ..., kw_n\} \\
\mathcal{KG}_{A}(E) &=& \{(v_i, v_j): v_i, v_j \in \mathcal{K}_A \text{ and } \\
& & v_i, v_j \in s_p, s_p \in S_A\}
\end{eqnarray*}

\subsubsection*{\textbf{Concept Detection for CIG Node Construction}} The $\mathcal{KG}$ captures the co-occurrence of keywords in sentences. However, since there could be multiple keywords in a long document, the vertices are highly specific and don't represent the primary themes/concepts in the document. The $\mathcal{KG}$ nodes are constructed at a fine granularity, while the CIG envisons nodes to represent major concepts in the document. Additionally, the $\mathcal{KG}$ is densely connected, and cannot be directly used as the concept interaction representation graph. To reduce the graph size, highly similar nodes in the $\mathcal{KG}$ can be clustered, and each cluster can represent a single node. To cluster highly correlated keywords, i.e., keywords that discuss similar concepts, community detection on the Keyword Graph $\mathcal{KG}$ is employed. $\mathcal{KG}$ contains densely connected components as keywords discussing a similar idea frequently co-occur in the same sentence. The community detection algorithm segregates keywords into clusters (or communities) by finding densely connected subgraphs. The individual cluster or community can be intuitively understood as a \emph{`Concept'}, such that concept $C^i = \{kw_m, kw_n, kw_o, ..., kw_p\}$ is a bag of keywords. The union of all concept communities of a document $\mathcal{D}_A$ is equal to the set of keywords $\mathcal{K}_A$. 
However, a keyword could belong to multiple communities or exclusively to just one community, depending on the community detection algorithm being used. Let us denote the set of concepts (or communities) for $\mathcal{D}_A$ as $\mathcal{C}_A$. 
\begin{align*}
\mathcal{C}_A = \bigcup_{i=1}^{p} C_{A}^{i} = \mathcal{K}_A = \mathcal{KG}_{A}(V) 
\end{align*}
As part of the ablation study, we experiment with various community detection algorithms, namely Girvan Newmann~\cite{girvan2002community}, Greedy Modularity~\cite{clauset2004finding}, Louvain~\cite{blondel2008fast}, and Leiden~\cite{traag2019louvain}. The results indicate that the community detection algorithm does not impact the performance on the \textsc{iFixit} dataset, i.e. all community detection algorithms perform comparably.

Each community is a bag of keywords. To construct the Concept Interaction Graph (CIG) of document $\mathcal{D}_A$, the vertex set $CIG_{A}(V)$ is initialized with the communities, i.e., CIG vertices are the communities detected in the keyword graph $\mathcal{KG}_A$. In addition, a dummy vertex $C^{\text{D}}$ is also added (details in the upcoming paragraph on Sentence Assignment).
\begin{align*}
    CIG_{A}(V) = \{C_{A}^{i}: C_{A}^{i} \in \mathcal{C}_A\} \cup \{C_{A}^{\text{D}}\}
\end{align*}

\subsubsection*{\textbf{Sentence Assignment}}~\label{ssec:sent_ass} Each node in the CIG is a concept, which is a bag of keywords. The bag of keywords is represented using a TF-IDF vector. Based on the TF-IDF vector similarity, each sentence in the document is assigned to concept nodes if the similarity is greater than a threshold $\text{sim}_{\text{thresh}}$. $\text{sim}_{\text{thresh}}$ is a predefined threshold used for filtering out low-quality sentence-concept matches. Sentences that are not assigned to any concept are matched to a dummy vertex $C^{\text{D}}$ in the graph. The sentence assignment function $\mathcal{F}$ for $CIG_A(V, E)$ of $\mathcal{D}_A$, assigns sentences $s_i \in S_A$ to a set of concept vertices $SA_i \subseteq \mathcal{C}_A$ or the dummy vertex $C_A^{\text{D}}$.
\begin{align*}
    \mathcal{F} &: S_A \to \mathcal{C}_A \cup C_A^{\text{D}} \\
    SA_i        &= \{C_A^j: 1 \leq j \leq |\mathcal{C}_A|, \\
                & \text{TFIDFSimilarity}(s_i, C_A^j) \ge \text{sim}_{\text{thresh}} \}
\end{align*}
$SA_i$ denotes the set of concepts that a sentence $s_i$ is assigned to based on the threshold. A sentence can be assigned to more than one concept. If the similarity of a sentence is less than the threshold similarity $\text{sim}_{\text{thresh}}$ with all the concept nodes, then it is assigned to the dummy vertex $C_A^{\text{D}}$.
\begin{equation*}
    \mathcal{F}(s_i) = 
    \begin{cases}
        SA_i, & SA_i \neq \phi \\
        C_A^{\text{D}}, & SA_i = \phi 
    \end{cases}
\end{equation*}

\subsubsection*{\textbf{Weighted Edge Computation}}
The Sentence Assignment stage assigns sentences to the concept vertices. Each vertex is represented by the TF-IDF vector of all the sentences assigned to it. Weighted edges of format (source, target, weight) are added to the CIG of document $\mathcal{D_A}$:
\begin{align*}
    CIG_{A}(E) = \{(v_i, v_j, w): v_i, v_j \in CIG(V), \\ i\neq j,  w=\text{TFIDFSim}(v_i, v_j)\}
\end{align*}
To summarize, keywords are extracted and clustered using community detection to form concepts, which form the vertices in the CIG. Based on the occurrence of keywords in sentences, each sentence is assigned to a concept node. Sentences assigned to a concept node are utilized to form node representations using TF-IDF vectors. Weighted edges are added to the graph by computing the similarity between TF-IDF node representations.

\subsection{Joint Concept Interaction Graph}
\label{ssec:jcig_met}
CIGs represent a document as an undirected weighted graph. A joint concept interaction graph, motivated by CIGs, represents a pair of documents by capturing the similarities between the documents on different topics. The vertex set of JCIG for a pair of documents $\mathcal{D}_A$ and $\mathcal{D}_B$, represented as $JCIG_{AB}(V)$, is the union of the vertices of $CIG_A$ and $CIG_B$. Similar to sentence assignment $\mathcal{F}$ in CIG construction, the sentences in documents $\mathcal{D}_A$ and $\mathcal{D}_B$ are assigned to the vertices of the $JCIG_{AB}$. Weighted edge construction also follows the process of similarity computation between TF-IDF vertex representations and edges with weights greater than a threshold $w_{\text{thesh}}$ are added to the JCIG. However, it should be noted that the TF-IDF node representations of vertices in the JCIG consist of sentences from both documents, unlike CIG, which has sentences only from one document. For JCIG, the formulations of the vertex set $JCIG_{AB}(V)$, edge set $JCIG_{AB}(E)$, and the sentence assignment function $\mathcal{F}$ are as follows:
\begin{alignat*}{2}
    & \mathcal{F}  : S_A \cup S_B \to &&\; \mathcal{C}_A \cup \mathcal{C}_B \cup C_{AB}^{D}\\
    & JCIG_{AB}(V) = && CIG_{A}(V) \cup CIG_{B}(V) \\
    & JCIG_{AB}(E) = && \{(v_i, v_j, w): \\
    &                && v_i, v_j \in JCIG_{AB}(V), i\neq j, \\
    &                && w=\text{TFIDFSimilarity}(v_i, v_j), \\
    &                && w \ge w_{\text{thesh}}\}
\end{alignat*}

JCIG vertices capture the important topics in the documents. Sentences assigned to concepts capture the sentences from each document that discuss the topic. However, the sequential nature of the document is destroyed in the graph. For certain domains, such as instruction manuals, recipes, legal documents, and compliance documents, the sequence of steps is crucial. To retain the sequential information, we propose adding directed edges to the graph. To this end, we propose our framework for encoding sequential information in JCIGs.

\section{Encoding Sequential Information}
\label{sec:encseqinfor}
Sequential information is crucial in certain domains of text such as recipe documents, instruction manuals, and compliance documents. Often these documents contain a sequence of steps/instructions/actions that necessarily need to be performed in the specified sequence.
For example, an instruction manual on replacing keys in a keyboard specifically mentions that the right hook should be unlocked first, followed by lifting of the left pin hook. The sequence is important, and the reverse procedure won't help, and in some cases could also lead to damage to the device. Hence, similarity computation of a pair of documents should consider the sequence of information flow in each document while reasoning if they are same or not. As we have observed, JCIG does not consider sequential flow of information and the information is lost in JCIG representation. To overcome this limitation, we propose our algorithm of adding directed edges that encode the sequential information. The process involves two components: (i) Direction Pseudograph (details in Section~\ref{sec:dirpseudograph}), and (ii) Dominant Direction Deduction (details in Section~\ref{sec:ddd}).

\section{Direction Pseudograph}
\label{sec:dirpseudograph}
Sentences are the most granular component of a document that can be interpreted individually and convey information about the document. 
Hence, to encode the directionality of information flow among sentences in  JCIG, we propose a simple solution that starts by constructing a pseudograph, which is a multigraph i.e. multiple edges exist between vertices. In this section, we will consider the construction of a pseudograph for a document that will add directed edges as a representative of the sequential flow. Consider that we initialize a pseudo concept graph, by adding all extracted concepts (communities found via community detection on the keyword graph) as vertices. Next, we add edges to the graph that encode the sequential information of the document. Our motivation is to capture the sequential flow of information between concepts. We start with the first and the second sentence in the document and find the concepts they have been assigned to. Suppose $CS(s_1)$ and $CS(s_2)$ denote the concepts sets that sentence 1 and sentence 2 have been assigned. Next, we add a directed unweighted edge from each concept $cs_i \in CS(s_1)$ to $cs_j \in CS(s_2)$ to denote the direction of information flow from $cs_i$ to $cs_j$. Then the same process is repeated for sentence 2 and sentence 3. Similarly, we add directed unweighted edges between concepts contained in each consecutive pair of sentences, which lead to multiple edges between a pair of nodes. The edge set of the psuedograph is defined as follows, where edge $e_{mn} = (m, n)$ denotes a directed unweighted edge from node $m$ to $n$, and $|S_A|$ is the set of sentences in $\mathcal{D}_A$:
\begin{alignat*}{2}
&Pseudograph(V) &&= JCIG(V) \\
&Pseudograph(E) &&= \{e_{mn} = (m, n): \\
& && m \in \text{CS}(s_i), n \in \text{CS}(s_{i+1}), \\
& && m \neq n, 1 \leq i \leq |S_A|-1\}
\end{alignat*}

\section{Dominant Direction Deduction for Sparse Directed Graphs}
\label{sec:ddd}
In this section, we discuss two algorithms for constructing sparse directed graphs. To this end, we employ the pseudograph and CIGs or JCIGs. In the previous step, we encoded the pairwise sequential information between concept nodes based on the sequence of sentences in the document. However, the pseudograph can be densely connected and edges between a pair of nodes could exist in both directions, rendering the addition of sequential information useless. Multiple sentences assigned to multiple concepts will lead to multiple directed edges between concept vertices, which necessitates the deduction of a dominant direction between a pair of nodes. Intuitively, the dominant direction between a node pair captures the information flow from one concept to another. We formulate the concept of dominant direction deduction ($D^3$), which finds the dominant flow of sequential information given multiple parallel and opposite edges exist between a pair of nodes in the direction pseudograph. 

\begin{figure*}[!htbp]
\centering
\begin{subfigure}{.48\textwidth}
  \centering
  \includegraphics[width=.48\textwidth]{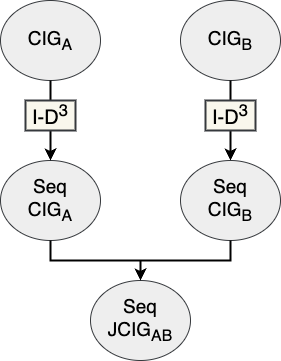}
  \caption{Individual - Dominant Direction Deduction (I-$D^3$)}
  \label{fig:id3}
\end{subfigure}%
\begin{subfigure}{.48\textwidth}
  \centering
  \includegraphics[width=.48\textwidth]{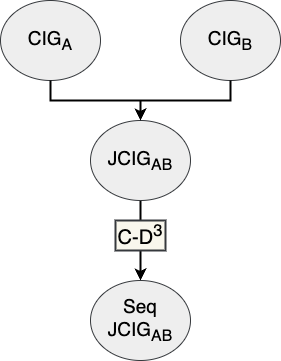}
  \caption{Combined - Dominant Direction Deduction (C-$D^3$)}
  \label{fig:cd3}
\end{subfigure}
\caption{Variants of the dominant direction deduction algorithm. Applying $D^3$ to a graph G, leads to a Seq G, a graph that encodes sequential information. There are two $D^3$ algorithms, SGS and HP, so overall, we propose four variants of $D^3$: I-SGS, C-SGS, I-HP, and C-HP.}
\label{fig:d3algos}
\end{figure*}

To reiterate, CIG is an undirected weighted graph for a single document and JCIG is an undirected weighted graph for a pair of documents. As we want to add directed edges from the psuedograph to CIG or JCIG to denote direction, we design two variants of determining dominant edges. The dominant direction deduction algorithm also leads to graph sparsification as multiple edges in the psuedograph between a pair of nodes in both direction, are reduced to just one edge in majority pairs. We represent the two variants of applying dominant direction deduction ($D^3$) algorithm in~\Cref{fig:d3algos}, (i) I-$D^3$: applying individually to the CIG, and (ii) C-$D^3$: applying on the combined JCIG of the document pair. Applying the variants with the two $D^3$ algorithms Supergenome Sorting (SGS) and Hamiltonian Path (HP), we discuss all four variants: Individual and Combined Supergenome Sorting (I-SGS and C-SGS respectively, in~\Cref{ssec:sgs}) and  Individual and Combined Hamiltonian Path (I-HP and C-HP respectively, in~\Cref{ssec:hp}). Our experiments reveal that C-HP performs the best.

\subsection{Supergenome Sorting (SGS)}
\label{ssec:sgs}
The graph resembles the supergenome graph presented in~\cite{gartner2018coordinate}, and we use a similar method for dominant direction deduction. The process starts by finding the pair of concept nodes in the graph (CIG in case individual $D^3$ is applied, or JCIG in case combined $D^3$) with the maximum number of edges, say concept $C^i$ and $C^j$. The number of edges from $C^i$ to $C^j$ and from $C^j$ to $C^i$ are counted. The final dominant direction between $C^i$ and $C^j$ corresponds to the direction with a greater number of edges. The same procedure is repeated for each pair of nodes in the pseudograph until the number of edges between a pair of nodes is reduced to one. In rare cases when the count of edges between a pair of nodes is same in both directions, a bidirectional edge is added.
To summarize, supergenome sorting algorithm takes the multigraph psuedograph and replaces multiple edges between a pair of nodes with weighted directed edges, with edge weight equal to the number of edges. Finally, the dominant direction between a pair of nodes corresponds to the direction which has the higher weight edge. The pseudograph with domainant directions received after this stage is then merged with the original CIG or JCIG by taking an intersection of the edges. The vertex set of pseudograph and CIG/JCIG is the same, i.e. concepts set. For the intersecting edges, the direction is adopted from the pseudograph and the weight of the edge is adopted from the CIG/JCIG, resulting in a sequential weighted directed graph (Seq $\text{CIG}_A$ or Seq $\text{JCIG}_{AB}$). In the worst case, $n^2$ edges in the original graph are reduced to $\frac{n^2}{2}$ after applying supergenome sorting.

Individual SGS (I-SGS) is applied to CIG of two documents $\mathcal{D}_A$ and $\mathcal{D}_B$ individually. I-SGS constructs the pseudographs (Pseudograph(A) and Pseudograph(B)) individually for the documents $\mathcal{D}_A$ and $\mathcal{D}_B$ from their CIGs, $\text{CIG}_A$ and $\text{CIG}_B$. I-SGS updates the edges in the pseudographs, by adding the dominant edges and removing other edges. After merging $\text{CIG}_A$ with Pseudograph(A) and $\text{CIG}_B$ with Pseudograph(B) (where pseudograph is updated with dominant directions), resulting in Seq $\text{CIG}_A$ and Seq $\text{CIG}_B$, which are finally combined to construct the Seq $\text{JCIG}_{AB}$ as discussed in~\Cref{ssec:jcig_met}. On the other hand, Combined SGS (C-SGS) constructs the Pseudograph(AB) from the $\text{JCIG}_AB$, applies the supergenome sorting on the joint psuedograph of document pairs, i.e. Pseudograph(AB), to get the sparse directed Pseudograph(AB). The resultant Pseudograph(AB) is intersected with $\text{JCIG}_AB$ resulting in a Seq $\text{JCIG}_AB$. It should be noted that the steps of supergenome sorting can be applied without any modification to the pseudograph of a CIG or JCIG as the steps are specific to a multigraph. The resulting pseudograph contains the dominant directions (in the form of a single directed edge between a pair of nodes), which is used to add directions to the final JCIG.

\subsection{Hamiltonian Path (HP)}
\label{ssec:hp}
A Hamiltonian Path in a graph is a path that visits each vertex only once. Given the pseudograph, which is a directed unweighted multigraph, each edge is assigned a weight. For an edge e = (m, n), denoting an edge from concept node m to concept node n, the weight is computed as the cosine similarity between the TF-IDF representations of concept m and concept n. Similar to the sentence assignment phase, each concept is represented by TF-IDF vectors based on the sentences that are assigned to the concept. Then the directed weighted multigraph is converted to a directed weighted simple graph by adding the weights of parallel edges between the same pair of nodes. Note that this is similar to constructing an edge with weight equivalent to the product of number of edges (in the same direction) between concept node pair and cosine similarity of concept nodes. It should be noted that the original pseudograph is densely connected and the resultant pseudograph after applying HP is a tournament graph in majority cases; and a hamiltonian path exists in a tournament graph. However, in cases the pseudograph is not a tournament graph, a simple addition of cosine similarity weighted edge between concept node pairs that don't have any edge, can convert the graph into a tournament graph. The resultant directed weighted simple pseudograph contains dominant directions, and is then merged with the CIG or JCIG in a similar way discussed in SGS in~\Cref{ssec:sgs}. 

In the worst case, $n^2$ edges in the original graph are reduced to $n$ edges after applying hamiltonian path algorithm. This sparisifies the graph while adding directed edges which encode sequential information. Sparsification leads to an advantage in the downstream architecture, where a GCN classifier operates on the graph, as we observe speedup in message passing as the number of edges is significantly reduced.

\subsection{Similarity Classification Architecture}
The JCIG encodes the information content from a pair of documents on concepts, which are major themes in the documents. Each concept node consists of sentences from both the documents if the concept occurs in both the documents. In case a concept occurs only in one document, then the node contains information about that concept from a single document only. We employ the architecture for document pair matching by~\cite{liu-etal-2019-matching}. The architecture consists of a Siamese encoder~\cite{neculoiu-etal-2016-learning} which learns match vectors for each vertex, followed by a graph convolution layer, node vector aggregation layer, and a multi-layer perceptron for binary classification. The schematic representation of the entire architecture is presented in~\Cref{fig:neural_architecture}. The graph nodes contain sentences for both documents, encoded via Word2Vec~\cite{mikolov2013efficient} and passed to the Siamese encoder. In case a concept does not contain sentences for a document, an average vector (average of all words in the vocabulary) is used to represent the document. Siamese encoder learns match vectors for each node by computing similarities and differences among sentences from both documents belonging to the concept. The learned match vector for each node is the node feature and the directed edges from our  $D^3$ algorithm gives the adjacency matrix of the graph. We use a three-layer GCN, and aggregate the hidden node representations using an aggregate layer that computes mean of hidden node vectors in the last layer. Finally, the mean of node vectors is passed to an MLP for binary classification, predicting if the document pairs are similar or not.
Let $c_{A}(v)$ and $c_{B}(v)$ denote the context vectors obtained via the siamese encoder for concept node $v$, for documents $\mathcal{D}_A$ and $\mathcal{D}_B$. The siamese encoder takes the Word2Vec representations of sentences from both documents, encodes the sentences assigned to node $v$ from each of the documents, and a match vector is constructed as follows, where $\circ$ denotes Hadamard product:
\begin{align*}
    m_{AB}(v) = (|c_{A}(v) - c{B}(v)|, c_{A}(v) \circ c_{B}(v))
\end{align*}

\section{Datasets and Experimental Setup}
\label{sec:data_exp}
We discuss the dataset details and the experimental setup in the upcoming subsections.
\subsection{CNSE, CNSS, and \textsc{iFixit} Dataset}
\label{ssec:data}
\textbf{CNSE and CNSS Dataset:} The Chinese News Same Event dataset (CNSE) and Chinese News Same Story dataset (CNSS) are curated by~\cite{liu-etal-2019-matching} to evaluate the performance of JCIG in document matching. The dataset contains news stories collected from news providers such as Sina, Sohu, Tencent, WeChat, etc., and the task is to identify stories discussing the same event or story. The dataset has a pair of news articles with binary labels denoting if the two articles are discussing the same event/story. CNSE and CNSS datasets are not sensitive to sequential information, and hence we did not initially propose to use our methodology for such documents. However, we evaluate our methodology on CNSE and CNSS dataset and show that performance does not deteriorate on these datasets in comparison to the JCIG baseline.

\noindent \textbf{\textsc{iFixit} Dataset:} We curate instruction manuals from the \href{https://www.ifixit.com/}{\textsc{iFixit}} website, which contains step-by-step guides in English for repair and assembly of appliances and devices across fifteen categories. In this work, we only curate manuals for the category `Appliances', consisting of 1333 manuals across twelve categories such as refrigerator, washer, water dispenser, kitchen appliances, etc. The dataset contains manuals that discuss the same procedure (such as assembly, disassembly, and replacement of a specific part) for appliances from different brands. For example, there are `water filter replacement' guides for Whirpool and Samsung refrigerators. Similarly, there are guides for other tasks such as screen replacement, thermal fuse replacement, condenser cleaning, etc. which are organized by different subjects. Each guide also has a brand tag, which indicates the brand of the appliance or device and tags about the components focused in the guide. To construct positive document pairs from the \textsc{iFixit} Appliances, we filter guides with the same subject and having at least two common tags. We construct negative document pairs by grouping documents that don't have the same subject.

\subsection{Experimental Setup}
\label{ssec:expsetup}
Our model is implemented using PyTorch and we follow standard 60-20-20 data splits for training, validation, and test set. We utilized 3 Quadro RTX 5000 GPU (16G memory) for running our experiments. We train all models for 50 epochs with early stopping. We run ablations with learning rates: 0.01, 0.005, 0.001, 0.0005, and 0.0001. We achieve the best results with a learning rate of 0.0005 across all configurations. We also experimented with 2, 3, and 4 GCN layers, and 3-layer GCN performed the best. We train word2vec using gensim library~\cite{rehurek_lrec}, and experiment with word vectors of sizes 100 and 200. All documents were tokenized with NLTK library before word2vec training. We use community detection algorithms available in NetworkX during concept detection stage.

\begin{table}[tbp]
\centering
\small{
    \begin{tabular}{c|cc|cc|cc}
    \hline
    \rowcolor{Gray}
    \multirow{2}{*}{\cellcolor{Gray}} \textbf{Method$\to$} & \multicolumn{2}{c}{\textbf{JCIG}} & \multicolumn{2}{c}{\textbf{SGS}} & \multicolumn{2}{c}{\textbf{HP}}  \\  \cline{2-7}
    \cellcolor{Gray} \textbf{Dataset$\downarrow$} & \textbf{Acc}    & \textbf{F1}    & \textbf{Acc}    & \textbf{F1}    & \textbf{Acc}   & \textbf{F1}    \\
    \hline
    \textbf{CNSE}  & 80.8 & 78.5 & 81.1 & 78.7 & \textbf{81.2}  & \textbf{79.0}  \\
    \textbf{CNSS}  & 86.4 & 86.7 & 86.3 & 86.6 & \textbf{86.6}  & \textbf{87.1}  \\ 
    \hdashline 
    \textbf{\textsc{iFixit}}  & 60.9 & 37.8 & 64.4 & 48.1 & \textbf{70.7}  & \textbf{58.5} \\  \hline
    \end{tabular}
}
\caption{Results on the CNSE, CNSS, and \textsc{iFixit} dataset. HP shows significant improvement in accuracy and F1 scores over the JCIG baseline for the \textsc{iFixit} dataset. We observe that HP performs superior to SGS in all cases. SGS and HP perform equally well on the CNSE and CNSS datasets, but significant improvements are not observed as sequential information is not crucial in news articles.}
\label{tab:results}
\end{table}
\section{Results and Observations}
\label{sec:results}
Even though CNSE and CNSS datasets don't contain any crucial sequential information, our proposed SGS and HP based dominant direction deduction methods achieve performance comparable to the JCIG baseline. The \textsc{iFixit} dataset contains instruction manuals for appliances and sequential information is critical. HP algorithm shows significant improvement in accuracy (\Cref{tab:results}) and achieves an improvement in accuracy of ten points in comparison to JCIG baseline. SGS also outperforms JCIG on the \textsc{iFixit} dataset by four points. During ablation, we observed that HP outperforms SGS and C-$D^3$ outperforms I-$D^3$.
\section{Conclusion}
Encoding sequence information in graph representations of document pairs is crucial for recipe documents, instruction manuals, and compliance documents. While JCIGs perform well for detecting similar news articles, we showcase that they perform poorly for instruction manuals in the \textsc{iFixit} dataset. We propose Supergenome sorting and Hamiltonian path algorithms for deducing dominant direction between graph nodes and also sparsify the graph by adding directed edges. Our proposed HP algorithm beats the JCIG baseline by ten points on the \textsc{iFixit} dataset. 

\section{Limitations}
The CNSE and CNSS datasets are Chinese news articles, while the \textsc{iFixit} dataset is in English. We posit that the performance of our methods will carry forward similarly to other languages. However, our method is limited by keyword detection algorithms and the availability of text-processing libraries in other languages which can significantly impact the downstream performance.

\section{Ethics Statement}
The curated dataset of \textsc{iFixit} manuals does not contain any information that is related to individuals, and neither does it contain any offensive content. The proposed method for document similarity has no ethical considerations.

\clearpage

\bibliographystyle{splncs04}
\bibliography{custom,anthology}

\end{document}